\documentclass{article}

\usepackage{arxiv}

\usepackage[utf8]{inputenc} 
\usepackage[T1]{fontenc}    
\usepackage{url}            
\usepackage{booktabs}       
\usepackage{amsfonts}       
\usepackage{nicefrac}       
\usepackage{microtype}      
\usepackage{lipsum}

\usepackage{geometry}
\usepackage{fancyhdr}
\usepackage{afterpage}
\usepackage{graphicx}
\usepackage{amsmath,amssymb,amsbsy}
\usepackage{dcolumn,array}
\usepackage{tocloft}
\usepackage{longtable}
\usepackage{mathtools}
\usepackage{todonotes}

\DeclareMathOperator*{\argmax}{argmax} 
\usepackage{comment}

\title{Template-based Question Answering using Recursive Neural Networks}

\author{
    Ram G Athreya \\
    Arizona State University \\
    \texttt{Ram.G.Athreya@asu.edu} \\
    \And
    Srividya Bansal \\
    Arizona State University \\
    \texttt{Srividya.Bansal@asu.edu} \\
    \And
    Axel-Cyrille Ngonga Ngomo \\
    Data Science Group, Paderborn University, Germany \\
    \texttt{axel.ngonga@upb.de} \\
    \And
    Ricardo Usbeck \\
    Data Science Group, Paderborn University, Germany, \\ Fraunhofer IAIS, Dresden, Germany \\
    \texttt{ricardo.usbeck@upb.de}
}

\begin{document}
\maketitle

\begin{abstract}
The Semantic Web contains large amounts of related information in the form of knowledge graphs (KGs) such as DBpedia or Wikidata. These KGs are typically enormous and are not easily accessible for users as they need specialized knowledge in query languages (such as SPARQL) as well as deep familiarity with the ontologies used by these KGs. To make these KGs more accessible (even for non-experts) several natural language question answering (QA) systems have been developed. Due to the complexity of the task, different methods have been tried including techniques from natural language processing (NLP), information retrieval (IR), machine learning (ML) and the Semantic Web (SW). Most question answering systems over KGs approach the question answering task as a conversion from the natural language question to its corresponding SPARQL query. 

This has lead to NLP pipeline architectures that integrate components that solve a specific aspect of the problem and pass on the results to subsequent components for further processing eg: DBpedia Spotlight~\cite{spotlight} for named entity recognition, RelMatch for relational mapping, etc. A major drawback of this approach is error propagation through the pipeline. 
Another approach is to use query templates either manually generated or extracted from existing benchmark datasets to generate the SPARQL queries. These templates are a set of predefined queries with various slots that need to be filled. This approach potentially shifts the question answering problem into a classification task where the system needs to match the input question to the appropriate template (class label).

We propose a neural network-based approach to automatically learn and classify natural language questions into its corresponding template using recursive neural networks. An obvious advantage of using neural networks is the elimination of the need for laborious feature engineering that can be cumbersome and error-prone. The input question is encoded into a vector representation. The model is trained and evaluated on the LC-QuAD dataset (Large-scale Complex Question Answering Dataset). The LC-QuAD queries are annotated based on 38 unique templates that the model attempts to classify. The resulting model is evaluated against both the LC-QuAD dataset and the 7th Question Answering Over Linked Data (QALD-7) dataset. The recursive neural network achieves template classification accuracy of 0.828 on the LC-QuAD dataset and an accuracy of 0.618 on the QALD-7 dataset. When the top-2 most likely templates were considered the model achieves an accuracy of 0.945 on the LC-QuAD dataset and 0.786 on the QALD-7 dataset. After slot filling, the overall system achieves a  macro F-score 0.419 on the LC-QuAD dataset and a macro F-score of 0.417 on the QALD-7 dataset.

\end{abstract}

\section{Introduction}
Knowledge graphs (KGs) are typically enormous and not easily accessible for users as they need specialized knowledge in query languages (SPARQL), as well as deep familiarity with the underlying ontologies. So, to make these KGs more accessible, several QA systems have been developed over the last decade. At a high level, most QA systems approach the task as a conversion from a natural language question to its corresponding SPARQL query using NLP pipelines. These systems then utilize the query to retrieve the desired entities or literals. Höffner et al.~\cite{qa-survey} classify the techniques used in QA systems over Linked Data broadly into five tasks: 

\begin{enumerate}
     
\item \textbf{Question Analysis:} The question of the user is analyzed based on purely syntactic features. QA systems use syntactic features to deduce, for example, the right segmentation of the question, determine the corresponding instance (subject or object), property or class and the dependency between different phrases.

\item \textbf{Phrase Mapping:} This step starts with a phrase (one or more words) $s$, and tries to find, in the underlying
KG, a set of resources that correspond to $s$ with high probability. $s$ could correspond to an instance, property or a class from the KG.

\item \textbf{Disambiguation:} Two ambiguity problems can arise. The first is that from the question analysis step the segmentation and the dependencies between the segments are ambiguous. For example, in the question "Give me all European countries" the segmentation can group or not group the expression "European countries" leading to two possibilities. Next, the phrase mapping step returns multiple possible resources for one phrase. In the example above "European" could map to different meanings of the word "Europe".

\item \textbf{Query Construction:} This phase deals with how the QA system constructs the SPARQL query to find the answer to the question. A problem arises during the query construction, that is commonly referred to as the "semantic gap". Assume for example that a user asks the question: "which countries are in the European Union?". Instead of a property $dbo:member$, DBpedia uses the class $dbc:Member\_states\_of\_the\_European\_Union$ to encode the information. The "semantic gap" refers to the problem that the KG encodes information differently from what one could deduce from the question. This shows that in general, it is difficult to deduce the form of the SPARQL query knowing only the question. 








\item \textbf{Querying:} The final step is to query the underlying KG to retrieve the answers for the given question. The answer can be from a single KG or depending on the system and the task even from multiple KGs.

\end{enumerate}

Error propagation in such pipelines can lead to crucial ramifications downstream and adversely affect the overall performance of the system. Error propagation becomes especially difficult for complex queries that span multiple triples, where many facts need to be discovered before the question can be answered. Current research follows two paths, namely (1) template-based approaches, that map input questions to either manually or automatically created SPARQL query templates or (2) template-free approaches that try to build SPARQL queries based on the given syntactic structure of the input question. However, template-free approaches require an additional effort of ensuring to cover every possible basic graph pattern, making it a more computationally intensive process~\cite{qa-challenges-survey}.

In this paper, we present \textbf{template classification as an alternative to the query building} approach or the sub-graph generation (from entities) approach. Furthermore, as the analysis of Singh et al.~\cite{frankenstein} on QALD subtasks shows, \textbf{query building has one of the poorest F-Measures} at 0.48. So, by performing template classification in the beginning, the workflow gets inverted and provides the benefit of restricting the number of resources, entities and ontology classes that need to be considered for a candidate SPARQL query instead of seemingly endless combinations, as is usually done in a non-template approach. In this article and for completeness, we focus on template classification only and use existing methods to fill the slots after the template classification to provide a performance comparison against existing methods. 
Our contributions are as follows:
\begin{itemize}
    \item We present a novel QA template classification model using recursive neural networks to replace the traditional query building process.
    \item Our approach can generalize to different domains/benchmark datasets. We showcase this by training on LC-QuAD only and testing it on the QALD-7 dataset. We emphasize that labeled data or large training data in the form of natural language question-SPARQL pairs is costly to (re-)produce.
    \item The resulting model was \textit{evaluated} using the FAIR GERBIL QA~\cite{gerbil_QA} framework resulting in 0.419 macro f-measure on LC-QuAD and 0.417 macro f-measure on QALD-7.
    \item The model was implemented using the Pytorch deep learning framework based on a well-known Tree-LSTM based on~\cite{rnn_lstm}\footnote{https://github.com/dasguptar/treelstm.pytorch}. Our model was adapted from this source code and is available online at \url{https://github.com/ram-g-athreya/RNN-Question-Answering} together with supplementary material.
\end{itemize}

\section{Related Work} \label{related_work}
Since the steady growth of the Semantic Web, the necessity for natural language interfaces to ontology-based repositories has become more acute, igniting interest in QA systems~\cite{qa-sw-useful}. In the last number of years, different complex benchmarks for QA systems over KGs have been developed. Most popular among them in the Semantic Web community is the \textbf{QALD} dataset~\cite{qald_9}. QALD is not one benchmark but a series of annual evaluation campaigns for QA systems with 9 iterations of the challenge to date. Another interesting dataset is the \textbf{LC-QuAD} dataset~\cite{lc_quad} that was developed from the ground up to facilitate machine learning based QA approaches using crowd workers. There is also DBNQA~\cite{hartmann-marx-soru-2018} which is a an offspring of LC-QuAD and QALD which does not offer new templates but new slots. Since we focus in this paper on template classification, we did not evaluate on DBNQA.

The key QA tasks in non-end-to-end systems comprise of Named Entity Recognition and Disambiguation, Relation Extraction and Query Building. No single system will be perfect for all tasks and across all domains~\cite{frankenstein}. 
This has led to the development of QA components that specialize in specific tasks for specific domains which can then be bootstrapped into modular pipelines.
The framework by Diefenbach et al.~\cite{qanary}, a message-driven and light-weight architecture, leverages linked data technology and particularly vocabularies to create a component-based QA system. Their RDF-based modular approach solves a critical problem in the community, that is, integrating existing components, which is a resource intensive process. 
The efficiency of these components was studied by training classifiers which take features of a question as input and have the goal of optimizing the selection of components based on those features ~\cite{frankenstein}. Then a greedy algorithm is used to identify the best pipeline that includes the best possible components which can effectively answer the given question. 
The system was evaluated using the QALD and LC-QuAD benchmarks where they discovered that among the available solutions for the three tasks in QA, Named Entity Recognition ranks the highest (based on Macro Precision, Recall and F-Score) followed by Query Building and finally Relation Linking. 
\textbf{WDAqua}~\cite{DBLP:conf/esws/DiefenbachSM17} is a monolithic rule-based system using a combinatorial approach to generate SPARQL queries from natural language questions, leveraging the semantics encoded in the underlying KG. It can answer questions on both DBpedia (supporting English) and Wikidata (supporting English, French, German and Italian). WDAqua does not require training and was also evaluated on QALD and LC-QuAD previously. Here, we reran the system as a baseline.
\textbf{ganswer2}~\cite{zou2014natural} is also a monolithic QA system which generates a semantic query graph, which reduced the transformation of a question to SPARQL to a subgraph matching problem. We also benchmarked against ganswer2.

The QA systems mentioned above translate questions into triples which are matched against an existing KG. However, in many cases, such triples do not accurately represent the semantic structure of the natural language question.
To circumvent this problem, Unger et al.~\cite{template-qa} proposed an approach that relies on a parse tree of the question to produce a SPARQL template that directly mirrors the internal structure of the question. This template contains empty slots which are then instantiated using statistical entity identification and predicate detection.
Lopez et al.~\cite{DBLP:conf/semweb/LopezTKW16} propose another template based QA system without the need to train a template classifier. The authors use the output of the dependency parse tree to create (linguistic) triples and identify the type of semantic entities.  Then they iterate a greedy algorithm for 2 to 3 rounds to determine the most similar template.
Abujabal et al.~\cite{Abujabal:2017:ATG:3038912.3052583} recently introduced an approach which can learn templates from user utterances. The templates are learned by distant supervision from question and Knowledge Graph answer pairs. The authors also employ dependency parse trees, which in turn allow leveraging compositional utterances. The templates are aligned between utterance and query by integer linear programming and learned in an offline step. In the online phase, the authors perform a light-weight template matching, consisting of automatically decomposing the question into constituent clauses and computing answers for each constituent using simple templates which are later combined to fully-fledged SPARQL queries. 

Since 2017, there is another line of work using neural networks for QA over KGs.
Soru et al.~\cite{DBLP:conf/i-semantics/SoruMMPVEN17} present a Neural SPARQL machine which is composed of three modules: a {generator}, a {learner}, and an {interpreter}.
The generator replaces placeholders in query templates with entities, properties, and labels, creating question-query pairs which are fed to the learner.
A bidirectional recurrent neural network based on LSTMs learns to parse questions and compose sequence representations of queries.
At prediction phase, the final query structure is then reconstructed by the interpreter through rule-based heuristics.
In a similar manner, Yin et al.~\cite{DBLP:journals/corr/abs-1906-09302} investigate eigth different neural machine translation architectures.
Their evaluation shows, that CNN-based architectures work better and that large, high-quality datasets are important as a foundation for effective models.

By contrast, in this paper recursive neural network would automatically learn the required representations through labeled examples provided in a large dataset, namely LC-QuAD. This methodology is domain independent and can thus be transposed to work with any domain requiring minimal additional modifications to the neural network architecture. 
Note, we used Tree-LSTMs in the domain of QA over Knowledge Graphs as a first step in this novel research directly and did not consider other baselines. Other baselines have been investigated by other works~\cite{rnn_lstm,DBLP:conf/semco/HakimovJC19}.

\section{LC-Quad Dataset}  

An essential requirement to develop and evaluate question answering systems is the availability of a large dataset comprising of varied questions and their corresponding logical forms. LC-QuAD consists of 5,000 questions along with the intended SPARQL queries required to answer questions over DBpedia. The dataset includes complex questions, i.e. questions in which the intended SPARQL query does not consist of a single triple pattern.

Trivedi et al.~\cite{lc_quad} generated the dataset by using a list of seed entities, and filtering by a predicate whitelist, generate subgraphs of DBpedia to instantiate SPARQL templates, thereby generating valid SPARQL queries. These SPARQL queries are then used to instantiate Normalized Natural Question Templates (NNQTs) which act as canonical structures and are often grammatically incorrect. These questions are manually corrected and paraphrased by reviewers. 

There are two key advantages for using LC-QuAD over similar existing datasets such as SimpleQuestions~\cite{simple_questions}, Free917n \cite{free917}, or QALD~\cite{qald_9}. They are:
\begin{enumerate}
\item Higher focus on complex questions unlike SimpleQuestions which focuses entirely on single triple patterns.
\item Larger volume and variety of questions. The Free917 dataset contains only 917 questions and QALD-9 has less than 1000 training and test questions combined.
\end{enumerate}

The LC-QuAD dataset contains 5,000 questions divided into 38 unique SPARQL templates comprising 5042 entities and 615 predicates. The SPARQL queries have been generated based on the 2016 DBpedia release. The dataset broadly contains three types of questions:
\begin{enumerate}
\item \textbf{Entity Queries:} Questions whose answer is an entity or list of entities with the WHERE clause containing one or more triples.
\item \textbf{Boolean Queries:} Questions whose answer is a boolean True or False with the WHERE clause containing exactly one triple.
\item \textbf{Count Queries:} Questions whose answer is a cardinal number with the WHERE clause containing one or more triples.
\end{enumerate}

Among the 5000 verbalized SPARQL queries, only 18\% are simple questions,
and the remaining queries either involve more than one triple, or COUNT/ASK keyword, or both. Moreover, 18.06\% queries contain a COUNT based aggregate, and 9.57\% are boolean queries. The advantage of using LC-QuAD is that it was tailored specifically for neural network approaches to question answering and has a relatively large variety of questions in the complex, count and boolean categories when compared to existing datasets which is valuable when training models and evaluating approaches. As of now, the dataset does not have queries with OPTIONAL, or UNION keywords. Also, it does not have conditional aggregates in the query head \cite{lc_quad}. 

Table \ref{table_lc_quad_distribution} tabulates the frequency distribution of each template in the LC-QuAD dataset along with its corresponding SPARQL template and an example query. Interestingly, the first 14 templates make up over 80\% of the dataset and there are 7 templates with under 10 examples. In fact, templates 601, 9 and 906 have only 1 example in the entire dataset.

\begin{longtable}{|l|l|p{1.5cm}|p{7cm}|p{3.5cm}|}
\hline
\multicolumn{1}{|r|}{\textbf{ID}} & \textbf{Count} & \textbf{Question Type} & \textbf{SPARQL Template} & \textbf{Example Query} \\ \hline

2 & 748 & Entity & SELECT DISTINCT ?uri WHERE \{ \textless r \textgreater \textless p \textgreater ?uri . \} & Name the mascot of Austin College ? \\ \hline

305 & 564 & Entity & SELECT DISTINCT ?uri WHERE \{ ?x \textless p \textgreater \textless r \textgreater . ?x \textless p2 \textgreater ?uri . ?x rdf:type \textless class \textgreater . \} & What layout can be found in cars similar to the Subaru Outback? \\ \hline

16 & 523 & Entity & SELECT DISTINCT ?uri WHERE \{ \textless r \textgreater \textless p \textgreater ?uri. \textless r2 \textgreater \textless p2 \textgreater ?uri . \} & Which series has an episode called The lost special and also a character named Sherlock Holmes ? \\ \hline

308 & 334 & Entity & SELECT DISTINCT ?uri WHERE \{?uri \textless p \textgreater \textless r \textgreater . ?uri \textless p2 \textgreater \textless r2 \textgreater . ?uri rdf:type \textless class \textgreater \} & Name the mountain whose range is Sierra Nevada (U.S.) and parent mountain peak is Nevado de Toluca? \\ \hline

301 & 309 & Entity &  SELECT DISTINCT ?uri WHERE \{ ?uri \textless p \textgreater \textless r \textgreater . ?uri rdf:type \textless class \textgreater \} & What is the river whose mouth is in deadsea? \\ \hline

3 & 262 & Entity & SELECT DISTINCT ?uri WHERE \{ \textless r \textgreater \textless p \textgreater ?x . ?x \textless p2 \textgreater ?uri . \} & What awards did the film director of The Haunted House win ? \\ \hline

5 & 213 & Entity & SELECT DISTINCT ?uri WHERE \{ ?x \textless p \textgreater \textless r \textgreater . ?x \textless p2 \textgreater ?uri . \} & Starwood operates in which places?	\\ \hline

15 & 198 & Entity & SELECT DISTINCT ?uri WHERE \{ \textless r \textgreater \textless p \textgreater ?uri. \textless r2 \textgreater \textless p \textgreater ?uri . \} & In which part of the world can i find Xynisteri and Mavro? \\ \hline

152 & 188 & Boolean & ASK WHERE \{ \textless r \textgreater \textless p \textgreater \textless r2 \textgreater . \} & Was Ganymede discovered by Galileo Galilei?	 \\ \hline

151 & 180 & Boolean & ASK WHERE \{ \textless r \textgreater \textless p \textgreater \textless r2 \textgreater . \} & Does the Toyota Verossa have the front engine design platform?	 \\ \hline

306 & 175 & Entity & SELECT DISTINCT ?uri WHERE \{ ?x \textless p \textgreater \textless r \textgreater . ?uri \textless p2 \textgreater ?x . ?uri rdf:type \textless class \textgreater \} & Which newspapers are owned by companies which are under Rolv Erik Ryssdal? \\ \hline

105 & 101 & Count & SELECT (COUNT(DISTINCT ?uri) as ?count) WHERE \{ ?x \textless p \textgreater \textless r \textgreater . ?x \textless p2 \textgreater ?uri . \} & How many awards have been given to screenwriters?	\\ \hline

1 & 159 & Entity & SELECT DISTINCT ?uri WHERE \{ ?uri \textless p \textgreater \textless r \textgreater . \} & What are the beverages whose origin is England?	\\ \hline

303 & 115 & Entity & SELECT DISTINCT ?uri WHERE \{ \textless r \textgreater \textless p \textgreater ?x . ?x \textless p2 \textgreater ?uri . ?x rdf:type \textless class \textgreater \} & What is the region of the ethnic group which speaks the language of Arkansas?	\\ \hline

6 & 94 & Entity & SELECT DISTINCT ?uri WHERE \{ ?x \textless p \textgreater \textless r \textgreater . ?uri \textless p2 \textgreater ?x . \} & What are some characters of the series produced by Ricky Grevais? \\ \hline

405 & 90 & COUNT & SELECT (COUNT(DISTINCT ?uri) as ?count) WHERE \{ ?x \textless p \textgreater \textless r \textgreater . ?x \textless p2 \textgreater ?uri . ?uri rdf:type \textless class \textgreater \} & How many companies have launched their rockets from the Vandenerg Air base?	\\ \hline

401 & 77 & Count & SELECT (COUNT(DISTINCT ?uri) as ?count) WHERE \{ ?uri \textless p \textgreater \textless r \textgreater . ?uri rdf:type \textless class \textgreater \} & How many places were ruled by Elizabeth II?	\\ \hline

111 & 76 & Count & SELECT (COUNT(DISTINCT ?uri) as ?count) WHERE \{ ?x \textless p \textgreater \textless r \textgreater . ?x \textless p \textgreater ?uri \} & Count the number of sports played by schools which play hockey ? \\ \hline

311 & 76 & Entity & SELECT ?uri WHERE \{ ?x \textless p \textgreater \textless r \textgreater . ?x \textless p \textgreater ?uri . ?x rdf:type \textless class \textgreater \} & Name all the doctoral student of the scientist who also supervised Mary Ainsworth ?	\\ \hline

406 & 70 & Count & SELECT (COUNT(DISTINCT ?uri) as ?count) WHERE \{ ?x \textless p \textgreater \textless r \textgreater . ?uri \textless p2 \textgreater ?x . ?uri rdf:type \textless class \textgreater \} & How many TV show has distributor located in Burbank California ?	\\ \hline

307 & 69 & Entity & SELECT DISTINCT ?uri WHERE \{ ?uri \textless p \textgreater \textless r \textgreater . ?uri \textless p \textgreater \textless r2 \textgreater . ?uri rdf:type \textless class \textgreater \} & What is the river that falls into North Sea and Thames Estuary? \\ \hline

101 & 67 & Count & SELECT (COUNT(DISTINCT ?uri) as ?count) WHERE \{ ?uri \textless p \textgreater \textless r \textgreater . \} & How many movies did Stanley Kubrick direct? \\ \hline

7 & 62 & Entity & SELECT DISTINCT ?uri WHERE \{ ?uri \textless p \textgreater \textless r \textgreater . ?uri \textless p \textgreater \textless r2 \textgreater \} & Whose former teams are Indianapolis Colts and Carolina Panthers?	\\ \hline

8 & 33 & Count & SELECT DISTINCT ?uri WHERE \{ ?uri \textless p \textgreater \textless r \textgreater . ?uri \textless p2 \textgreater \textless r2 \textgreater . \} & Which colonel consort is Dolley Madison? \\ \hline

102 & 26 & Count & SELECT (COUNT(DISTINCT ?uri) as ?count) WHERE \{ \textless r \textgreater \textless p \textgreater ?uri \} & How many states does the Pioneer corporation operate in?	\\ \hline

106 & 22 & Count & SELECT (COUNT(DISTINCT ?uri) as ?count) WHERE \{ ?x \textless p \textgreater \textless r \textgreater . ?uri \textless p2 \textgreater ?x . \} & Count all those whose youth club was managed by Luis Enrique. \\ \hline

11 & 20 & Entity & SELECT ?uri WHERE \{ ?x \textless p \textgreater \textless r \textgreater . ?x \textless p \textgreater ?uri . \} & List the outflows of the lake which has Benu river as one of it ? \\ \hline

403 & 17 & Count & SELECT (COUNT(DISTINCT ?uri) as ?count) WHERE \{ \textless r \textgreater \textless p \textgreater ?x . ?x \textless p2 \textgreater ?uri . ?x rdf:type \textless class \textgreater \} & How many countries surround the sea into which the Upper Neratva flow? \\ \hline

103 & 17 & Count & SELECT (COUNT(DISTINCT ?uri) as ?count) WHERE \{ \textless r \textgreater \textless p \textgreater ?x . ?x \textless p2 \textgreater ?uri . \} & How many other important things have been written by the creator of Stuart Alan Jones?	\\ \hline

108 & 14 & Count & SELECT (COUNT(DISTINCT ?uri) as ?count) WHERE \{ ?uri \textless p \textgreater \textless r \textgreater . ?uri \textless p2 \textgreater \textless r2 \textgreater . \} & How many bacteria have taxonomy as Bacillales and domain as Bacteria?	 \\ \hline

315 & 10 & Entity & SELECT DISTINCT ?uri WHERE \{ \textless r \textgreater \textless p \textgreater ?uri. \textless r2 \textgreater \textless p \textgreater ?uri .  ?uri rdf:type \textless class \textgreater \} & Which city is the resting place of the Martin Ragaway and Chuck Connors ?	\\ \hline

402 & 9 & Count & SELECT (COUNT(DISTINCT ?uri) as ?count) WHERE \{ \textless r \textgreater \textless p \textgreater ?uri . ?uri rdf:type \textless class \textgreater \} & How many teams was Garry Unger in, previously?	\\ \hline

316 & 5 & Entity & SELECT DISTINCT ?uri WHERE \{ \textless r \textgreater \textless p \textgreater ?uri . \textless r2 \textgreater \textless p2 \textgreater ?uri . ?x rdf:type \textless class \textgreater \}  & List the people casted in Betsy's Wedding and 16 candles?	\\ \hline

107 & 5 & Count & SELECT DISTINCT COUNT(?uri) WHERE \{ ?uri \textless p \textgreater \textless r \textgreater . ?uri \textless p \textgreater \textless r2 \textgreater . \} & Count the number of shows whose creators are Jerry Seinfeld and Larry David?	 \\ \hline

605 & 2 & Entity & SELECT DISTINCT ?uri WHERE \{ ?x \textless p \textgreater \textless r \textgreater . ?x \textless p2 \textgreater ?uri . ?x rdf:type \textless class \textgreater \} & What are the kind of games one can play on windows? \\ \hline

601 & 1 & Entity & SELECT DISTINCT ?uri WHERE \{ ?uri \textless p \textgreater \textless r \textgreater . ?uri rdf:type \textless class \textgreater \} & Which technological products were manufactured by Foxconn?	 \\ \hline

9 & 1 & Entity & SELECT DISTINCT ?uri WHERE \{ \textless r \textgreater \textless p \textgreater ?x . ?x \textless p \textgreater ?uri . \} & Who is owner of the soccer club which owns the Cobham Training Centre?	 \\ \hline

906 & 1 & Entity & SELECT DISTINCT ?uri WHERE \{ ?x \textless p \textgreater \textless r \textgreater . ?uri \textless p2 \textgreater ?x . ?uri rdf:type \textless class \textgreater \} & Name some TV shows whose theme is made by a band associated with Buckethead?	 \\ \hline
\caption{Frequency Distribution of Templates in LC-QuAD Dataset}
\label{table_lc_quad_distribution} \\
\end{longtable}

As shown in Table \ref{table_lc_quad_distribution} from the previous section there is great imbalance between the distribution of templates in the dataset. Also, some templates are exact replicas of others with an additional triple. For example, templates below 100 and templates in the 3xx series and templates in the 1xx and 4xx series have only one triple differentiating them: 

\begin{center}
    \textbf{?var rdf:type \textless class\textgreater }    
\end{center}

With this in mind, during preprocessing all templates which had less than 50 examples in the initial dataset were removed. The rationale here was that each template should have at least a 1\% representation in the final dataset. Also, templates below 100 were merged with their corresponding 3xx templates and 1xx templates were merged with 4xx templates by adding additional OPTIONAL queries to the SPARQL template. Also, templates 151 and 152 were merged into each other since they have identical SPARQL templates.
\newpage
For example template 1 and template 301 were combined into a single template as follows:

Template 1: \textbf{SELECT DISTINCT ?uri \{ ?uri \textless p \textgreater \textless r \textgreater . \}}

Template 301: \textbf{SELECT DISTINCT ?uri \{ ?uri \textless p \textgreater \textless r \textgreater . ?uri rdf:type \textless class \textgreater \}}

Combined Template: \textbf{SELECT DISTINCT ?uri \{ ?uri \textless p \textgreater \textless r \textgreater . OPTIONAL \{ ?uri rdf:type \textless class \textgreater \} \}}

The removal of sparse templates resulted in only 80 questions being removed and the final dataset had 4,920 questions spread across 15 templates. The frequency distribution and updated templates of the  preprocessed dataset are shown in Table \ref{table_preprocessed_frequency}. It must be noted that this refined dataset was used to train the template classification model. In spite of the manual review process there were several grammatical mistakes and misspellings of proper nouns in the dataset which were corrected as needed and the results of the same is shared with the LC-QuAD team so that they can improve the quality of the dataset for the community.

\begin{longtable}{|l|p{2cm}|l|p{1.75cm}|p{9.25cm}|}
\hline
\multicolumn{1}{|r|}{\textbf{ID}} & \textbf{Templates Merged} & \textbf{Count} & \textbf{Question Type} & \textbf{New SPARQL Template} \\ \hline

5 & 5, 305 & 777 & Entity & SELECT DISTINCT ?uri WHERE \{ ?x \textless p \textgreater \textless r \textgreater . ?x \textless p2 \textgreater ?uri . OPTIONAL \{ ?x rdf:type \textless class \textgreater \} \} \\ \hline

2 & 2 & 748 & Entity & SELECT DISTINCT ?uri WHERE \{ \textless r \textgreater \textless p \textgreater ?uri . \} \\ \hline

16 & 16 & 523 & Entity & SELECT DISTINCT ?uri WHERE \{ \textless r \textgreater \textless p \textgreater ?uri . \textless r2 \textgreater \textless p2 \textgreater ?uri . \} \\ \hline

1 & 1, 301 & 468 & Entity & SELECT DISTINCT ?uri WHERE \{ ?uri \textless p \textgreater \textless r \textgreater . OPTIONAL \{ ?uri rdf:type \textless class \textgreater \} \} \\ \hline

3 & 3, 303 & 377 & Entity & SELECT DISTINCT ?uri WHERE \{ \textless r \textgreater \textless p \textgreater ?x . ?x \textless p2 \textgreater ?uri . OPTIONAL \{ ?x rdf:type \textless class \textgreater \} \} \\ \hline

151 & 151, 152 & 368 & Boolean & ASK WHERE \{ \textless r \textgreater \textless p \textgreater \textless r2 \textgreater . \} \\ \hline

8 & 308 & 334 & Entity & SELECT DISTINCT ?uri WHERE \{ ?uri \textless p \textgreater \textless r \textgreater . ?uri \textless p2 \textgreater \textless r2 \textgreater . ?uri rdf:type \textless class \textgreater \} \\ \hline

6 & 6, 306 & 269 & Entity & SELECT DISTINCT ?uri WHERE \{ ?x \textless p \textgreater \textless r \textgreater . ?uri \textless p2 \textgreater ?x . OPTIONAL \{ ?uri rdf:type \textless class \textgreater \} \} \\ \hline

105 & 105, 405 & 261 & Count & SELECT (COUNT(DISTINCT ?uri) as ?count) WHERE \{ ?x \textless p \textgreater \textless r \textgreater . ?x \textless p2 \textgreater ?uri . OPTIONAL \{ ?uri rdf:type \textless class \textgreater \} \} \\ \hline

15 & 15 & 198 & Entity & SELECT DISTINCT ?uri WHERE \{ \textless r \textgreater \textless p \textgreater ?uri. \textless r2 \textgreater \textless p \textgreater ?uri . \} \\ \hline

101 & 101, 401 & 144 & Count & SELECT (COUNT(DISTINCT ?uri) as ?count) WHERE \{ ?uri \textless p \textgreater \textless r \textgreater . OPTIONAL \{ ?uri rdf:type \textless class \textgreater \} \} \\ \hline

7 & 7, 307 & 131 & Entity & SELECT DISTINCT ?uri WHERE \{ ?uri \textless p \textgreater \textless r \textgreater . ?uri \textless p \textgreater \textless r2 \textgreater . OPTIONAL \{ ?uri rdf:type \textless class \textgreater \} \} \\ \hline

111 & 111 & 76 & Count & SELECT (COUNT(DISTINCT ?uri) as ?count) WHERE \{ ?x \textless p \textgreater \textless r \textgreater . ?x \textless p \textgreater ?uri \} \\ \hline

11 & 311 & 76 & Entity & SELECT ?uri WHERE \{ ?x \textless p \textgreater \textless r \textgreater . ?x \textless p \textgreater ?uri . ?x rdf:type \textless class \textgreater \} \\ \hline

106 & 406 & 70 & Count & SELECT (COUNT(DISTINCT ?uri) as ?count) WHERE \{ ?x \textless p \textgreater \textless r \textgreater . ?uri \textless p2 \textgreater ?x . ?uri rdf:type \textless class \textgreater \} \\ \hline
\caption{Frequency of templates after preprocessing. Templates with  50 examples removed and similar templates merged}
\label{table_preprocessed_frequency} \\
\end{longtable}
\section{Template Classification Approach} \label{approach}
Our proposed system follows the steps mentioned below:(i) Question Analysis; (ii) Template Classification (Query Construction); (iii) Slot-Filling (Phrase Mapping and Disambiguation); (iv) Querying. 
The first two steps are presented in this section and remaining two steps are elaborated in Section~\ref{slot_filling}. This is because the output from step 1 is directly used in step 2 and the same is true for steps 3 and 4.

\subsection{Question Analysis}
First, the question provided by the user is analyzed based on purely syntactic features. QA systems use syntactic features to deduce, for example, the right segmentation of the question, determine which phrase corresponds to an instance, property or class and the dependency between the different phrases~\cite{qa-survey}. For now, we only deal with syntactic parsing of the incoming question in this phase and converting it into a form that can be used for training the Recursive Neural Network. 

\subsubsection{Part of Speech Tagging}
Part-of-Speech (POS) Tagging is the process of annotating a word in a text as corresponding to a particular part of speech,e.g.: noun, verb, adjective, etc. In Natural Language Processing (NLP) applications, POS tagging is usually the first step in a pipeline and the output of POS tagging is typically used by downstream processes such as parsing for instance.

For the model, the English version of the Stanford POS tagger was used~\cite{pos}. The Stanford POS Tagger is a log-linear POS tagger which utlilizes both preceding and following tag contexts through the implementation of a dependency network representation. The tagger uses the Penn Treebank Tagset \cite{penn_treebank} for tagging the individual parts of speech and the Java implementation (v3.9.1) of the tagger was used.

\begin{figure}[htb!]
\centering
    \includegraphics[width=0.8\linewidth]{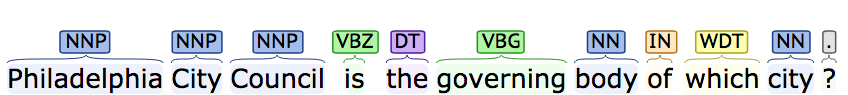}
    \caption{Stanford POS-Tagger Output}
    \label{pos_output}
\end{figure}

For example, consider the question \textit{"Philadelphia City Council is the governing body of which city ?"}. The corresponding POS tagged question is represented in Figure~\ref{pos_output}.

\subsubsection{Dependency Parsing}
Parsing in NLP is the process of determining the syntactic structure of text  using a formal grammar. Given a sentence, a parser computes the combination of production rules that generate the sentence according to the underlying grammar. POS tagged information alone is not enough to identify the relationships between the different chunks in a question. But this information can be leveraged by parsers to provide rich meaningful information between constituent words. The Stanford Neural Network dependency parser was used by the system \cite{parser}. The input to the parser was the sequence of POS tags generated from the previous step and the output is the corresponding parse tree. The Java implementation (v3.9.1) on the Stanford parser was used by the system.

\begin{figure}[htb!]
\centering
    \includegraphics[width=0.8\linewidth]{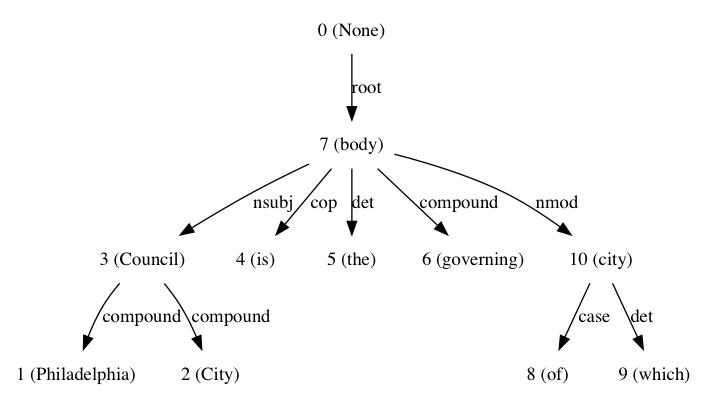}
    \caption{Stanford Dependency Parser Output}
    \label{dep_tree_output}
\end{figure}

Figure~\ref{dep_tree_output} represents the Stanford Dependency Parser output for the question \textit{"Philadelphia City Council is the governing body of which city ?"}.

There are two outputs of interest from Dependency Parsing. The first is the typed dependencies for each word in the input sentence. The typed dependencies representation provide a simple description of the grammatical relationships in a sentence. Its expressed as triples of a relation between pairs of words. For the rest of the paper these relationships are denoted as RELS. The second output is the parse tree. 




\subsection{Input Preparation}
The output from the parse tree needs to be vectorized so that they can be supplied to the neural network model. There are two strategies to vectorize words: 1) One-Hot Encoding or 2) Word Embeddings.  
Based on the data, five distinct kind of input models were developed for training. These are summarized in Table~\ref{table_input_models}.

\subsubsection{One-Hot Encoding}
One-Hot encoding is a common strategy in machine learning for converting categorical input into a vector by setting all values as 0 except for 1 bit which has a value 1, hence the name One-Hot. For example the number of POS tags in the LC-QuAD dataset is 43. So each POS tag is represented as a 43 x 1 vector where a single index is 1 and the rest are 0 depending on the index of the POS tag being considered. This conversion from an abstract categorical value to a consistently sized vector enables easier processing and prediction by machine learning models.

\subsubsection{Word Embedding}
Generally, the goal of word embeddings is mapping the words in unlabeled text data to a continuously-valued low dimensional space, in order to capture the internal semantic and syntactic information. The concept of word embedding was first introduced with the Neural Networks Language Model (NNLM). They are usually unsupervised models and incorporate various architectures such as Restricted Boltzmann Machine (RBM), Convolutional Neural Network (CNN), Recurrent Neural Network (RNN) and Long-Short Term Memory (LSTM) that can be used to build word embeddings. Usually the goal of the NNLM is to maximize or minimize the function of log likelihood, sometimes with additional constraints \cite{word_embedding}. A key reason for using word embedding is that, in the past few years it has been shown that pre-trained models produce vastly better performance compared to existing methods such as one-hot vectors.

\subsection{Facebook FastText}
For word embedding the system uses Facebook's FastText embedding model \cite{fasttext}. FastText uses an approach based on the skipgram model (taking into account subword information), where each word is represented as a bag of character n-grams. The main advantage of using FastText is its ability to handle out of vocabulary words better. The dataset had over 6000 unique tokens which were compressed into word vectors of dimensions 300 x 1 using the FastText word embedding model.

\begin{table}[htp]
    \begin{tabular}{|c|c|p{8cm}|}
    \hline
    \textbf{Model} & \textbf{Dimensionality} & \textbf{Description} \\ \hline
    POS & 43 x 1 & Only POS Tags expressed as One-Hot Vector \\ \hline
    POS + RELS & 85 x 1 & One-Hot POS vector concatenated with One-Hot RELS\footnote{RELS denote the relationships derived from the Dependency Parse of all questions. eg: nsubj, pobj, etc. There were 42 unique RELS tags in the dataset.} Vector \\ \hline
    FastText & 300 x 1 & FastText Word Embedding \\ \hline
    FastText + POS + RELS & 385 x 1 & FastText Word Vector concatenated with One-Hot POS and RELS Vector \\ \hline 
    FastText + POS + RELS + CHARS & 444 x 1 & FastText Word Vector concatenated with One-Hot POS, RELS and CHARS\footnote{The CHARS vector for a word is the average of One-Hot Vectors of the characters of each word in the question} Vector \\ \hline
    \end{tabular}
    \caption{Dimensionality of different models created for the template classification task}
    \label{table_input_models}
\end{table}

\subsection{Recursive Neural Network}
A recursive neural network is basically an extension of a recurrent neural network implemented on a graph or tree-based input instead of a sequential input. 
They are non-linear adaptive models that are able to learn deep structured information. They were introduced as promising machine learning models for processing data from structured domains. They can be employed for both classification and regression problems and are capable of solving both supervised and unsupervised tasks. They provide the flexibility of being able to work with input of arbitrary length compared to other feature based approaches which are constrained to fixed length vectors \cite{rrn_intro}. 

Here, the Tree-LSTM was implemented based on the model proposed by Tai et al.~\cite{rnn_lstm} and our architecture is based on their implementation. Tree-LSTM is a generalization of LSTMs to tree-structured network topologies. A key distinction between Tree-LSTM and standard LSTM is that, while the standard LSTM composes its hidden state from the input at the current time step and the hidden state of the LSTM unit in the previous time step, the tree-structured LSTM, or Tree-LSTM, composes its state from an input vector and the hidden states of arbitrarily many child units. The standard LSTM can then be considered a special case of the Tree-LSTM where each internal node has exactly one child. 

Similar to standard LSTM units, each Tree-LSTM unit (indexed by $j$) contains input and output gates $i_j$ and $o_j$, a memory cell $c_j$, hidden state $h_j$ and input vector $x_j$ where $x_j$ is a vector representation of a word in a sentence. The critical difference between the standard LSTM unit and Tree-LSTM units is that gating vectors and memory cell updates for a given node are dependent on the states of its child units. Additionally, instead of a single forget gate, the Tree-LSTM unit contains one forget gate $f_{jk}$ for each child ${k}$. This allows the Tree-LSTM unit to selectively incorporate information from each child. For example, a Tree-LSTM model can learn to emphasize semantic heads in a semantic relatedness task, or it can learn to preserve the representation of sentiment-rich children for sentiment classification \cite{rnn_lstm}.

Given a tree, let $C(j)$ denote the set of children of node $j$. The Tree-LSTM transition equations are the following:
\begin{equation}
    \widetilde{h_j} = \sum_{k \epsilon C(j)} h_k
\end{equation}

\begin{equation}
    i_j = \sigma (W\textsuperscript{(i)}x_j + U\textsuperscript{(i)}\widetilde{h_j} + b\textsuperscript{(i)}) 
\end{equation}

\begin{equation}
    f_{jk} = \sigma (W\textsuperscript{(f)}x_j + U\textsuperscript{(f)}h_k + b\textsuperscript{(f)})
\end{equation}

\begin{equation}
    o_j = \sigma (W\textsuperscript{(o)}x_j + U\textsuperscript{(o)}\widetilde{h_j} + b\textsuperscript{(o)})
\end{equation}

\begin{equation}
    u_j = \sigma (W\textsuperscript{(u)}x_j + U\textsuperscript{(u)}\widetilde{h_j} + b\textsuperscript{(u)})
\end{equation}

\begin{equation}
    c_j = i_j \odot u_j + \sum_{k \epsilon C(j)} f_{jk} \odot c_k
\end{equation}

\begin{equation}
    h_j = o_j \odot tanh(c_j)
\end{equation}

The Tree-LSTM learns a question by passing the sequence of words and the tree structure. Although the tree begins at the root, the model recursively traverses the tree and first learns the hidden states of the leaf nodes. The state of the leaf nodes are used by their corresponding parents to derive their state and so on until the network finally reaches the root node. So learning occurs breadth first from the leaf to the root. Finally, the output from the root node is converted into a $N_t$ dimensional vector using a softmax classifier where $N_t$ is the number of templates which in this case is 15. Formally, to predict template $\hat{t}$ from the set of templates $T$ we calculate the softmax at the root node followed by the argmax to classify the template for the given question as shown below:

\begin{equation}
    \begin{multlined}
        \hat{p}_\theta( t | {x}_{root} ) = softmax(W\textsuperscript{(s)}h_{root} + b\textsuperscript{(s)}), \\
        \hat{t} = \argmax_t \hat{p}_\theta( t | {x}_{root} )
    \end{multlined}
\end{equation}

The cost function is the negative log-likelihood of the true class label $y$ and $\lambda$ is the L2-Regularization hyperparameter as given below:
\begin{equation}
    J(\theta) = - log\hat{p}_\theta( y | {x_{root}} ) + \frac{\lambda}{2}||\theta||_2^2
\end{equation}

\section{Slot Filling Approach} \label{slot_filling}
For a given input question, the template classification algorithm from the previous section determines the top-n (in our case $n=2$ to omit computing overhead) templates that are most likely to answer the question. The template captures the semantic structure of the user's query, which is then mapped to the underlying knowledge graph, leaving gaps only for the slots that need to be injected as needed. The candidate SPARQL template broadly contains three kinds of slots that need to be filled:
\begin{enumerate}
    \item \textbf{Resources:} are named entities (proper nouns), which can be detected using standard entity recognition tools. For example London, Microsoft, etc.
    \item \textbf{Predicates:} are nouns, adjectives, or verbs that may modify a resource. For example: born, capital, etc.
    \item \textbf{Ontology Classes:} Ontology classes that are associated with resources define the type of class a resource might fall under. For example, when considering the resource Barack Obama (dbr:Barack\_Obama) a valid ontology class would be Person (dbo:Person). Ontology classes are linked through the \textit{rdf:type} predicate of the target resource.
\end{enumerate}

For example, consider the question \textit{"Philadelphia City Council is the governing body of which city ?"}. The underlying candidate template detected for this question form would be: 

\noindent\texttt{SELECT DISTINCT ?uri \{ ?uri p r. OPTIONAL \{ ?uri rdf:type class\}\}}

As can be seen, for answering this question, one resource, one predicate and one ontology class need to be detected. It must be noted that the ontology class detection is optional, and even though the original candidate SPARQL query from the LC-QuAD dataset does not require an ontology class, the present system requires it since the LC-QuAD templates 1 \& 301 were merged during the data preprocessing step, see Section~3. 
An ensemble of tools was used for the slot filling process. The reason for using multiple tools for a given task was to cover the weaknesses of each while at the same time maximizing their strengths to produce the best possible results. Note, we do not focus on the slot filling part in this paper. 

For named entity recognition, DBpedia Spotlight~\cite{spotlight} and TagMe~\cite{tagme} were used. DBpedia Spotlight automatically annotates text with DBpedia URIs, aka resources. For the slot filling task, a confidence of 0.4 (default) was used. The specialty of TagMe is that it may annotate texts that are short and poorly composed to underlying Wikipedia pages and their inter-relations. Singh et al.~\cite{frankenstein} showed that TagMe outperforms other Named Entity Recognition tools on the LC-QuAD dataset and hence it was a natural choice for this task. But TagMe suffers when it comes to the detection of single word entities, such as Geneva (dbr:Geneva) in the question: \textit{"Is Esther Alder the mayor of Geneva?"}. But DBpedia Spotlight has better accuracy in spotting short entities while struggling with multi-word entities, which are detected more efficiently by TagMe. These are hence a good complementary solution. Wherever TagMe detected multi-word entities, these were ranked higher compared to the entities detected by DBpedia Spotlight.

For relation and class linking, Singh et al.~\cite{frankenstein} state that RNLIWOD\footnote{https://github.com/dice-group/NLIWOD} has the best overall performance on the LC-QuAD dataset but their results also show that it has poor overall macro performance (0.25 precision, 0.22 recall \& 0.23 F-1 score). 
Thus, we augmented its dictionary of predicates and ontology classes along with their \textit{rdfs:label} used in the DBpedia Chatbot project~\cite{chatbot}. This resulted in higher coverage of predicates and classes that could be matched with the input question, thereby leading to better performance. The lexicon is a key-value hashmap with the keys being the various surface forms that can be used to express a particular predicate or class and their value being all possible predicates or classes that match the sequence of words in the given surface form.

After the candidates for each slot are detected, candidate queries are built using the Cartesian product of the possible values in each slot. Each combination is queried against a DBpedia 2016-10 SPARQL endpoint, which was the latest stable release compatible with LC-QuAD and QALD at the time of writing, to determine if they yield any results. This process continues until the first viable combination is discovered that produces results against the endpoint. As Usbeck et al~\cite{hawk} showed, the problem of SPARQL query generation and pruning of invalid candidate queries is very computationally intensive and very little progress has been made beyond the semantic analysis of the Cartesian product approach to improve both efficiency and performance in this part of the QA process. 

\newpage
\section{Experimental Results}\label{results}
In this section, we present the used model parameters and experimental results, followed by a discussion of the findings.

\subsection{Model Selection \& Hyperparameter Tuning}
\begin{figure}[htb!]
    \centering
    \includegraphics[width=0.8\linewidth]{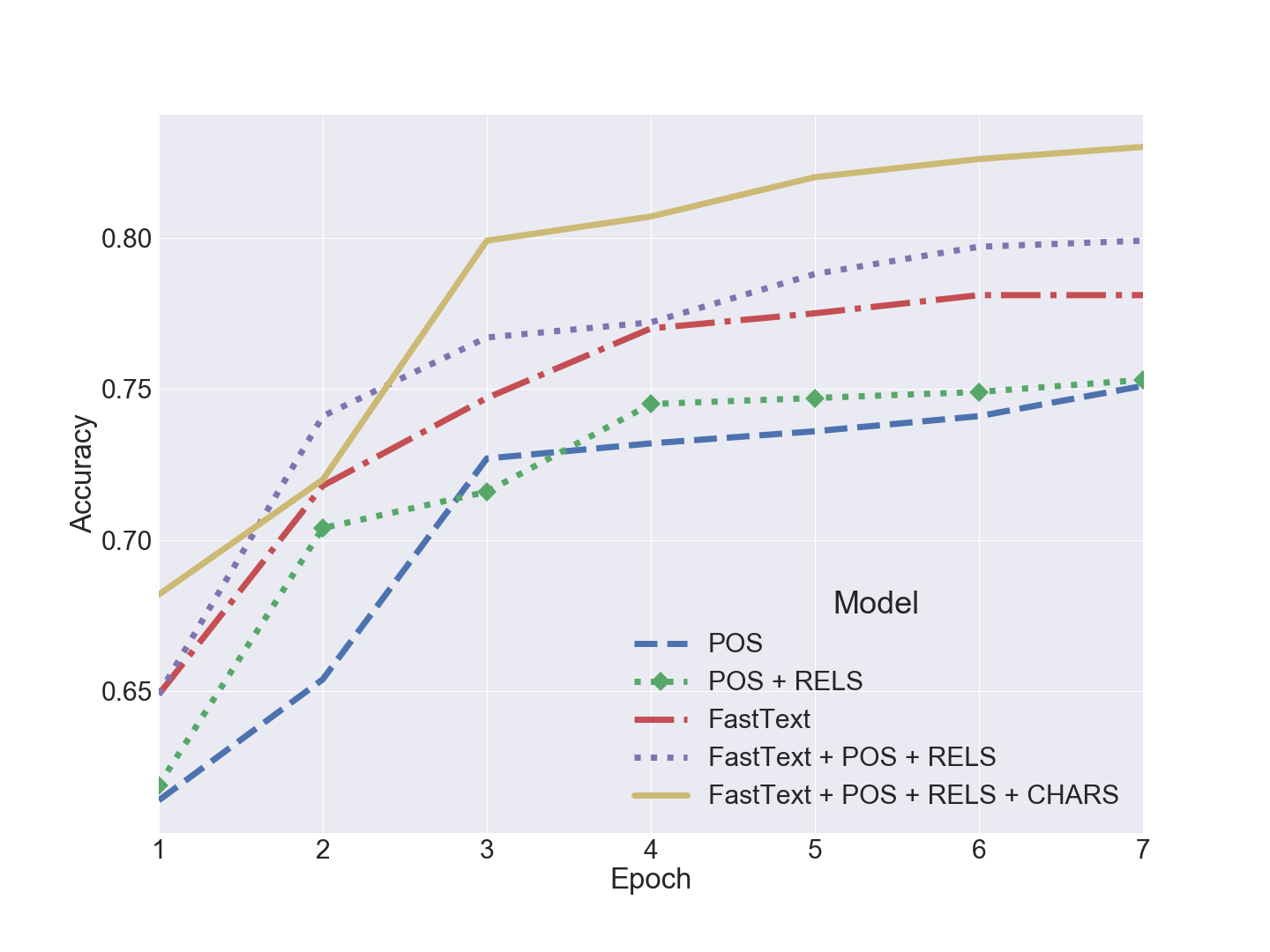}
    \caption{Accuracy on the test dataset for different input models}
    \label{model_accuracy}
\end{figure}

Among the different models for input that were attempted, the model that produced the best results was the one that used a combination of FastText Word Embedding concatenated with the One-Hot Vectors of the POS tag and word dependency relationship (RELS) derived from the syntactic parse of the sentence combined with the average of the One-Hot character vectors of each character in a given word. Figure~\ref{model_accuracy} shows the accuracy across epochs for each of the model combinations that were considered and clearly shows that the FastText + POS + RELS + CHARS model outperforms all other combinations of input. The preprocessed dataset containing 4920 questions was split into train and test datasets with a split of 80\% training and 20\% test data. The accuracy of this model was 0.828 on the test dataset. We calculated accuracy as  $ accuracy(y, \hat{y}) = \frac{1}{N} \sum_{i=1}^{N} 1 (\hat{y_i} = y_i) $, where $\hat{y_i}$ is the predicted value of the $i\-th$ example, $y$ is the corresponding true value and $N$ is the total number of examples. 

Table~\ref{table_model_parameters} tabulates the hyperparameters of the model. The input vector was the concatenated 444-dimensional word vector. The Adam Optimizer~\cite{adam} was used with a mini batch size of 25 examples. The loss function used was Cross - Entropy Loss, which has been shown to exhibit superior performance for tasks involving multivariate classification~\cite{loss_survey}. Due to the low number of training examples, the model had to be aggressively regularized and the learning rate periodically curtailed to prevent overfitting while simultaneously improving the model's generalization performance. Three strategies were employed to achieve this: (1) Weight Decay, (2) Dropout, (3) Adaptive Learning Rate.

\begin{table}[htb!]
    \centering
    \begin{tabular}{lr}
    \toprule
        \textbf{Parameter} & \textbf{Value} \\ 
        \midrule
        Input Dimensions & 444 x 1 \\ 
        LSTM Memory Dimensions & 150 x 1 \\ 
        Epochs & 7 \\ 
        Mini Batch Size & 25 \\ 
        Learning Rate & 1 x 10\textsuperscript{-2} \\ 
        Weight Decay (Regularization) & 2.25 x 10\textsuperscript{-3} \\ 
        Embedded Learning Rate & 1 x 10\textsuperscript{-2} \\ 
        Dropout & 0.2 \\ 
        Loss Function & Cross - Entropy Loss \\ 
        Optimizer & Adam Optimizer \\ 
        Learning Rate Scheduler & Stepwise Learning Rate Decay \\ 
        Step LR Step Size & Once every 2 epochs \\ 
        Step LR Decay & 0.25 \\ 
        \bottomrule
    \end{tabular}
    \caption{Model Parameters for our RNN model.}
    \label{table_model_parameters}
\end{table}

\begin{figure}[htb!]
    \centering
    \includegraphics[width=0.9\linewidth]{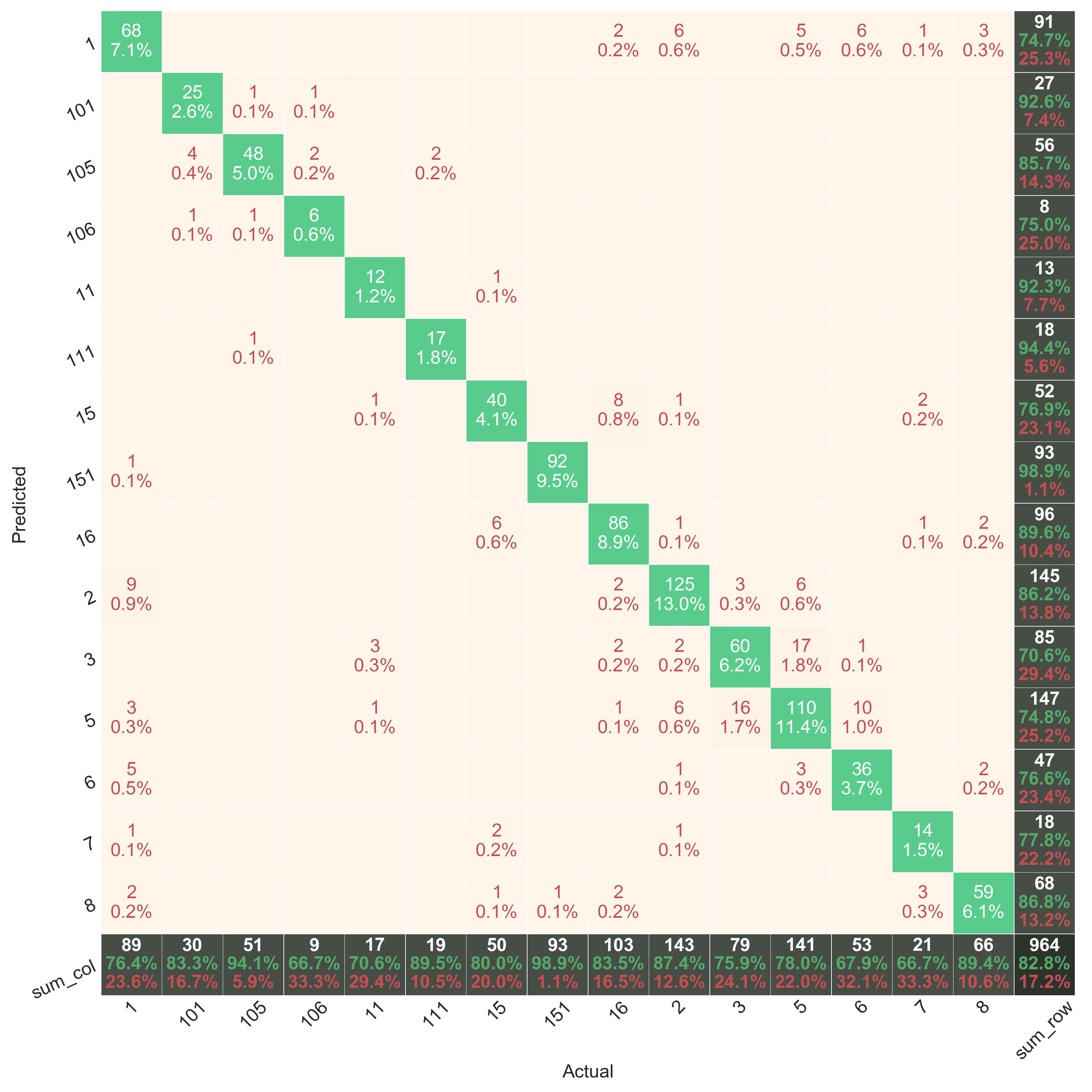}
    \caption{Confusion Matrix}
    \label{confusion_matrix}
\end{figure}

\subsection{Template Classification} \label{section_template_classification_results}
The best model from the template classification task produced an accuracy of 0.828 and  0.945 in the top-2 templates. Table~\ref{table_lc_quad_accuracy} displays template level accuracy. The number of examples does not seem to affect the accuracy at the template level. Rather, based on the confusion matrix from Figure~\ref{confusion_matrix} it can be observed that specific templates misclassify each other. For example, templates 3 \& 5 are more likely to misclassify each other, and the same can be said for 5 \& 6 but 3 \& 6 do not misclassify. Also, template 1 misclassifies with template 2 at a much higher rate since they are mirrors of a single triple pattern. That is, template 1 has the triple pattern \texttt{?uri  p   r } while template 2 has the triple pattern \texttt{r   p  ?uri}. To test how well the model generalizes it was also tested on the QALD-7~\cite{qald7} multilingual dataset without any additional training or optimizations. The model had never seen the dataset before and hence can serve as a good candidate to test the model's predictive power on never before seen data. The test dataset was not considered due to several issues, e.g., unseen namespaces such as Dublin Core. We were not able to use the entire QALD-7 dataset because we trained the model on LC-QuAD which does not contain examples outside the DBpedia ontology schema. Note, we trained on the whole LC-QuAD dataset as other (unpublished) approaches\footnote{\url{https://github.com/AskNowQA/KrantikariQA}} did, as there is no dedicated development set. The training dataset of QALD-7~\cite{qald7} contains 215 questions, of which 85 examples were eliminated during pre-processing. The model was tested on a total of 130 examples which is roughly ~60\% of the dataset and represented 7 templates that were analogous in the LC-QuAD dataset. These 7 templates are a subset of the 15 templates from LC-QuAD. The remaining questions were manually tagged by us based on the similarity of their SPARQL queries to the LC-QuAD dataset. The reasons why questions were eliminated are as follows:
\begin{enumerate}
\item \textbf{Filter \& Union based queries:} As already mentioned, the LC-QuAD dataset currently does not support FILTER, OPTIONAL or UNION queries which do feature in the QALD dataset. \item \textbf{MinMax Queries:} MinMax queries as the name suggests are natural language questions that ask for a variation of minimum or maximum of something eg: highest, lowest, largest, smallest, longest, shortest, etc. 
\item \textbf{Many Triples:} Some questions require 3 or more triples to answer. 
\item \textbf{Complex Boolean Questions:} Currently LC-QuAD's Boolean questions have only a single triple in the where clause. In contrast, the QALD dataset also contains examples of questions with 2 triples and several variations of complex queries for boolean questions which LC-QuAD does not support. 
\end{enumerate}

\begin{table}[htb!]
    \centering
    \begin{tabular}{@{}rrrr@{}}
        \toprule
        \textbf{Template} & \textbf{\#Examples} & \textbf{Acc.} & \textbf{Top-2 Acc.} \\ 
        \midrule
        2 & 80 & 0.68 & 0.84 \\ 
        1 & 18 & 0.66 & 0.94 \\ 
        151 & 12 & 1.0 & 1.0 \\ 
        3 & 12 & 0.25 & 0.42 \\ 
        8 & 6 & 0.00 & 0.33 \\ 
        5 & 1 & 0.00 & 0.00 \\ 
        11 & 1 & 0.00 & 0.00 \\ 
        \bottomrule
    \end{tabular}
    \caption{Template Level Model Accuracy on the QALD dataset.}
    \label{table_qald_accuracy}
\end{table}

\begin{table}[htb!]
    \centering
    \begin{tabular}{@{}rrr@{}}
        \toprule
        \textbf{Template} & \textbf{\#Examples} & \textbf{Accuracy} \\
        \midrule
        2 & 143 & 0.87  \\ 
        5 & 141 & 0.78 \\ 
        16 & 103 & 0.83 \\ 
        151 & 93 & 0.98 \\ 
        1 & 89 & 0.76 \\ 
        3 & 79 & 0.75 \\ 
        8 & 66 & 0.89 \\ 
        6 & 53 & 0.67 \\ 
        105 & 51 & 0.94 \\ 
        15 & 50 & 0.80 \\ 
        101 & 30 & 0.83 \\ 
        7 & 21 & 0.66 \\ 
        111 & 19 & 0.89 \\ 
        11 & 17 & 0.70 \\ 
        106 & 9 & 0.66 \\
        \bottomrule
    \end{tabular}
    \caption{Template Level Model Accuracy on LC-QuAD dataset.}
    \label{table_lc_quad_accuracy}
\end{table}

Table~\ref{table_qald_accuracy} shows the template distribution breakdown for accuracy in the QALD dataset. The overall accuracy was 0.618 and the top-2 accuracy was 0.786. The performance varies considerably per dataset. This is because the quality of questions differs across datasets. Quality has various dimensions, such as complexity or expressiveness. Template 2 is over-represented compared to other templates, with some templates such as template 5 and template 11 having only 1 example. But the top 3 templates (by number of examples), which comprises ~84\% of the dataset, have a high top-2 accuracy, which shows reasonable generalization power for the template classification model. Although the original LC-QuAD dataset had 15 different targets while the QALD dataset had only 7, this did not contribute to a significant loss in accuracy since there was sufficient separation between disparate templates / question types, as shown in the confusion matrix in Fig \ref{confusion_matrix}. That is, while templates that were very similar to each other, such as template 1 and 2 (simple queries), tended to have a higher chance of misclassification between one another, they did not misclassify with template 151 (boolean query).

\begin{figure}[htb!]
    \centering
    \includegraphics[width=0.7\linewidth]{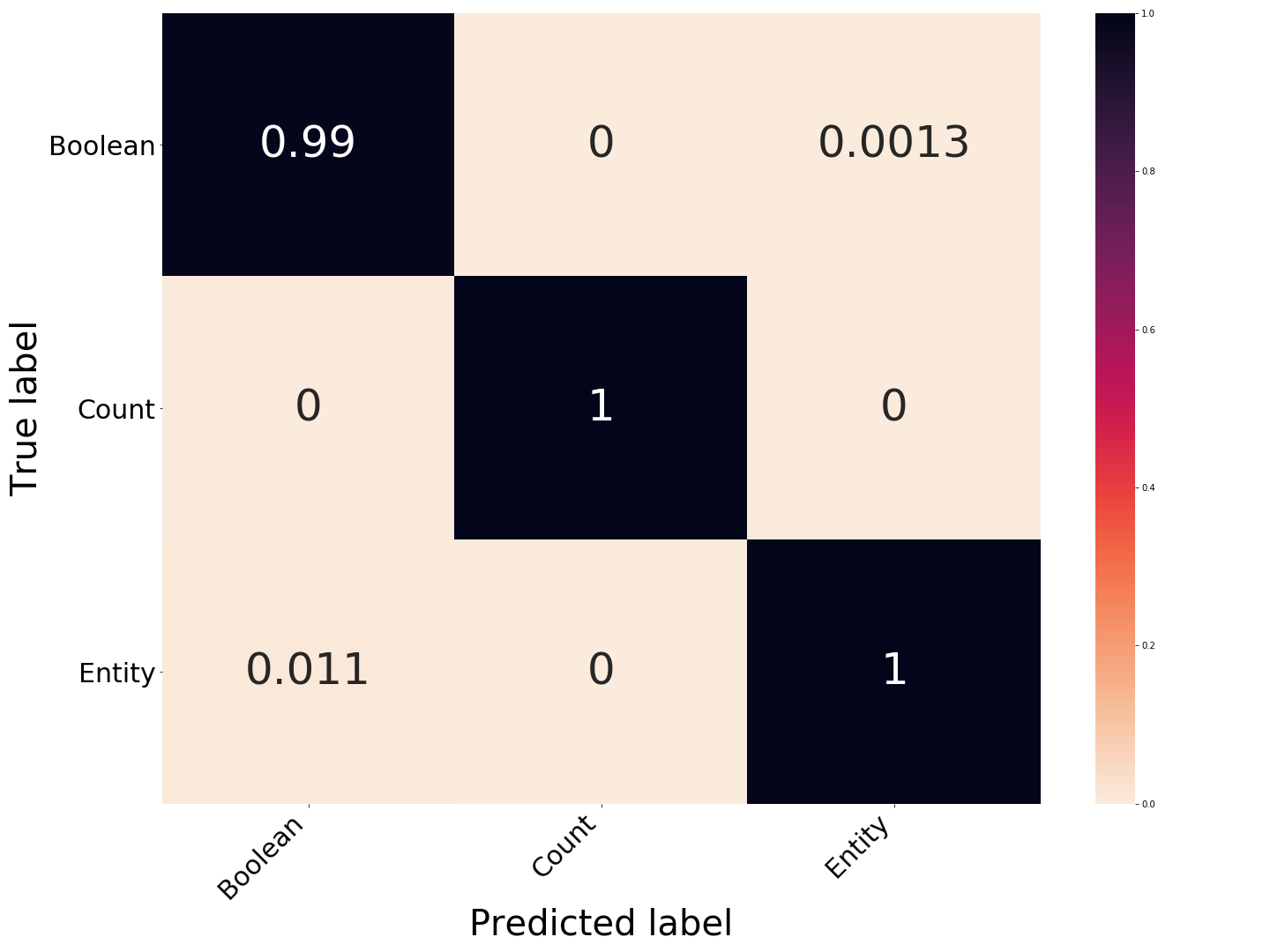}
    \caption{Answer Type Detection visualisation via a confusion matrix.}
    \label{answer_type_detection}
\end{figure}

An interesting byproduct of the model is its answer type detection capability, i.e., entity, count or boolean questions get efficiently grouped. 
The results of which are shown in Figure~\ref{answer_type_detection}.

\subsection{Slot Filling}
Table~\ref{table_lc_quad_f_scores} shows the performance of the system on the LC-QuAD test dataset. Table~\ref{table_qald7_f_scores} shows the performance of the system QALD-7 training dataset along with a comparison of the latest QA systems benchmarked on that dataset. We used internal methods of the GERBIL QA~\cite{gerbil_QA} framework to assist in testing. Although we do not outperform the state of the art in every case, we wanted to highlight a novel research avenue with this work. 

\begin{table}[htb!]
    \centering
    \begin{tabular}{p{3cm}p{1.5cm}p{1cm}p{1cm}p{1cm}}
        \toprule
        \textbf{LC-QuAD Test} & \textbf{Ontology Class} & \textbf{Predicate} & \textbf{Resource} & \textbf{Overall} \\ 
        \midrule
        Micro Precision & 0.802 & 0.950 & 0.976 & 0.135 \\ 
        Micro Recall & 0.150 & 0.178 & 0.206 & 0.064\\ 
        Micro F-1 Measure & 0.253 & 0.300 & 0.341 & 0.087 \\ 
        \midrule
        Macro Precision & 0.218 & 0.266 & 0.271 & 0.416 \\ 
        Macro Recall & 0.215 & 0.258 & 0.261 & 0.428 \\ 
        Macro F-1 Measure & 0.216 & 0.260 & 0.264 & 0.419 \\
        \bottomrule
    \end{tabular}
    \caption{Performance of system on LC-QuAD}
    \label{table_lc_quad_f_scores}
\end{table}
\noindent Reasons for errors in the named entity recognition task were:
 
\noindent \textbf{1. Specific instance detection:} Sometimes a specific form of an entity gets detected instead of the generic variety. Consider the question: \textit{"How many schools have bison as a mascot ?"}. The entity American Bison (dbr:American\_bison) was annotated instead of the generic bison (dbr:Bison).\\
\noindent \textbf{2. Disambiguation:} Sometimes it was hard to figure out the right entity to map to the resource when there were partial matches between the sequence of words in the question and the label of the corresponding entity. Consider the question: \textit{"Was 2658 Gingerich discovered at Harvard ?"}. Even though Harvard University (dbr:Harvard\_University) has a higher PageRank in the DBpedia knowledge graph and would be the correct choice for most questions, in this particular case the correct entity is Harvard College (dbr:Harvard\_College). \\
\noindent \textbf{3. Accented Characters:} Entities with foreign characters were detected poorly by both entity recognition tools. eg: Étienne Biéler  (dbr:Étienne\_Biéler). \\
\noindent \textbf{4. Colloquialisms:} Colloquial forms referring to well known entities were hard to detect. Consider the question \textit{"How many companies were started in the states ?"} the phrase "the states" refers to USA (dbr:United\_States) but instead State (Political) (dbr:State\_(polity)) was detected. 

\begin{table}[htb!]
    \centering
    \begin{tabular}{p{3cm}llp{1.75cm}}
        \toprule
        \textbf{QALD-7 Train} & \textbf{WDAqua} & \textbf{ganswer2} & \textbf{Proposed System} \\ 
        \midrule
        Micro Precision & - & 0.113 & 0.757 \\ 
        Micro Recall & - & 0.561 & 0.466 \\ 
        Micro F-1 Measure & - & 0.189 & 0.577 \\ 
        \midrule
        Macro Precision & 0.490 & 0.557 & 0.416 \\ 
        Macro Recall & 0.54 & 0.592 & 0.423 \\ 
        Macro F-1 Measure & 0.510 & 0.556 & 0.417 \\ 
        \bottomrule
    \end{tabular}
    \caption{Performance comparison on QALD-7~\cite{qald7}}
    \label{table_qald7_f_scores}
\end{table}

\noindent Reasons for errors in the relation extraction task were: 

\noindent \textbf{1. Implicit Predicates:} Sometimes the predicate needed to answer the question cannot be inferred from the question. Consider the question \textit{"How many golf players are there in Arizona State Sun Devils ?"} and its SPARQL query (Template 101): \texttt{SELECT COUNT(?uri) \{ ?uri  dbo:college dbr:Arizona \_State\_Sun\_Devils . ?uri rdf:type dbo:GolfPlayer \}}. 
To answer the question the predicate college (dbo:college) needs to be detected, but this is impossible to do with existing methods based on just the input question alone. \\
\noindent \textbf{2. Abbreviations:} Abbreviations instead of their expanded form were harder for relation linking tools to detect. Eg: PM for Prime Minister (dbo:primeMinister). \\
\noindent \textbf{3. Disambiguation:} The issue of disambiguation also plagues relation linking. The question, \textit{"What is the label of Double Diamond (album) ?"} refers to a record label (dbo:recordLabel), which was hard for the system to detect. \\
\noindent \textbf{4. Subset predicates:} Sometimes specific forms of a predicate needed to be detected e.g., head coach (dbp:headCoach) instead of coach (dbp:coach).

We did not compare ourselves to other works, since they either have a lower overall performance~\cite{frankenstein}, used a non-reproducible subset of LC-QuAD or have significantly changed their codebase since  publication~\cite{DBLP:conf/jist/AsakuraKYTT18,10.1007/978-3-319-93417-4_46,DBLP:conf/esws/ZiminaNJPSH18}.
\section{Conclusions \& Future Work}
This paper presents a novel approach for the QA over Linked Data task by converting it into a template classification task followed by a slot filling task. Although earlier template-based approaches have attempted similar solutions, this was the first time (to the best of our knowledge) that recursive neural networks were applied to the template classification task. For completeness, a slot filling approach using an ensemble of the best components for named entity, predicate and class recognition tasks were presented.
We answered the following research questions: 
\begin{enumerate}
    \item Can state-of-the-art neural network techniques such as Long Short Term Memory (LSTM), recursive neural networks, and word embeddings be leveraged for the template classification task? \\ Yes, our evaluation showed that the template classification model achieved an accuracy of 0.828 accuracy and 0.945 top-2 accuracy on the LC-QuAD dataset and an accuracy of 0.6183 and 0.786 top-2 accuracy on the QALD-7 dataset. After slot filling the system achieves a macro F-score 0.419 on the LC-QuAD dataset and a macro F-score of 0.417 on the QALD-7 dataset.
    \item Can a template classification model serve as a replacement for the query building process that has been shown to be both error-prone and computationally intensive \cite{frankenstein,hawk,qald_f_score}? \\ Yes, our model can address the template classification task without the need for expensive feature engineering. 
    \item Can the template classification model be developed without any domain specific information/features that can make it easily transferable? \\ Yes, our template classification model was developed without any domain specific information or features as long as it is QALD-formatted. Thus it can easily be transferred across domains using appropriate, KB-agnostic slot filling tools~\cite{DBLP:conf/esws/MoussallemURN18}.
\end{enumerate}

\noindent We are aware that our approach has a coverage issue in terms of being bound to the training templates and look forward to mitigating this issue through a finer-grained training process. While the choice of LSTMs seems arbitrary and does not outperform the state of the art in all respects, we aimed to provide a proof-of-concept for a domain-agnostic QA system. Basing a domain-agnostic QA system on template classification alleviates the need for costly feature engineering that is characteristic of classical machine.  We refer to Hakimov et al.'s~\cite{DBLP:conf/semco/HakimovJC19} intuition that until now there is no systematic way to explore neural network architectures for a specific task. Thus, exploring other neural networks is certainly a possible research direction. For instance, we will explore an encoder setting on top of the input words instead of using pretrained embedding.

The new insight we gained about the pairwise misclassification of specific templates points to a potential future research direction where this recursive neural network model can serve as a drop-in.
Also, the template classification approach can be extended to predict only certain segments of the final SPARQL query~\cite{Abujabal:2017:ATG:3038912.3052583}. 
For domain adoption, the templates can either be reused or will need to be constructed from new training data. We are aware that the existence of particular templates limits the types of the queries a system can handle, and will strive to remedy this issue by template decomposition in upcoming research.

\bibliographystyle{unsrt}  
\bibliography{references}  

\end{document}